\documentclass[sigconf,screen]{acmart}

\AtBeginDocument{%
  \providecommand\BibTeX{{%
    \normalfont B\kern-0.5em{\scshape i\kern-0.25em b}\kern-0.8em\TeX}}}

\acmConference[]{}{}{}

\renewcommand\footnotetextcopyrightpermission[1]{}

\usepackage[utf8]{inputenc}
\usepackage{multirow}
\usepackage{xcolor}
\usepackage{pifont}

\begin{document}

\title{DiffHarmony: Latent Diffusion Model Meets Image Harmonization}

\author{Pengfei Zhou}
\affiliation{
 \institution{Beijing University of Posts \& Telecommunications}
 \state{Beijing}
 \country{China}
}
\email{zhoupengfei@bupt.edu.cn}

\author{Fangxiang Feng}
\authornote{Corresponding Author.}
\affiliation{
 \institution{Beijing University of Posts \& Telecommunications}
 \state{Beijing}
 \country{China}
}
\email{fxfeng@bupt.edu.cn}

\author{Xiaojie Wang}
\affiliation{
 \institution{Beijing University of Posts \& Telecommunications}
 \state{Beijing}
 \country{China}
}
\email{xjwang@bupt.edu.cn}

\begin{abstract}
Image harmonization, which involves adjusting the foreground of a composite image to attain a unified visual consistency with the background, can be conceptualized as an image-to-image translation task. Diffusion models have recently promoted the rapid development of image-to-image translation tasks . However, training diffusion models from scratch is computationally intensive. Fine-tuning pre-trained latent diffusion models entails dealing with the reconstruction error induced by the image compression autoencoder, making it unsuitable for image generation tasks that involve pixel-level evaluation metrics. To deal with these issues, in this paper, we first adapt a pre-trained latent diffusion model to the image harmonization task to generate the harmonious but potentially blurry initial images. Then we implement two strategies: utilizing higher-resolution images during inference and incorporating an additional refinement stage, to further enhance the clarity of the initially harmonized images. Extensive experiments on iHarmony4 datasets demonstrate the superiority of our proposed method. The code and model will be made publicly available at \href{https://github.com/nicecv/DiffHarmony}{https://github.com/nicecv/DiffHarmony}.
\end{abstract}



\maketitle

\section{Introduction}

Image composition faces a notable hurdle in achieving a realistic output, as the foreground and background elements may exhibit substantial differences in appearance due to various factors such as brightness and contrast. To address this challenge, image harmonization techniques can be employed to ensure visual consistency.  In essence, image harmonization entails refining the appearance of the foreground region to align seamlessly with the background. The rapid advancements in deep learning approaches ~\cite{xue2012understanding,zhu2015learning,tsai2017deep,cong2020dovenet,jiang2021ssh,guo2021image,sofiiuk2021foreground,Xi.Li.Xi.ea.2022,Ca.Sh.Ga.ea.2023,Ch.Gu.Li.ea.2023,Ha.Ya.Zh.ea.2023,Ta.Li.Ni.ea.2023} have contributed significantly to the progress of the image harmonization task. 

The input for the image harmonization task consists of a composite image and a foreground mask used to distinguish between the foreground and background, with the output being a harmonized image. In other words, both the input and output of the image harmonization task are in image format. Therefore, it can be viewed as an image-to-image translation task. Recently, diffusion models \cite{So.We.Ma.ea.2015,Di.Ti.Be.ea.2021,ho2020denoising} have significantly advanced the progress of image-to-image translation tasks. 
For instance, Chitwan \emph{et al.} \cite{Sa.Ch.Ch.ea.2022} proposed Palette, which is a conditional diffusion model that establishes a new SoTA on four image-to-image translation tasks, namely colorization, inpainting, uncropping, and JPEG restoration. Hshmat \emph{et al.} \cite{Hs.Da.Ch.ea.2023} proposed SR3+, which is a diffusion-based model that achieves SoTA results on blind super-resolution task. 

Directly applying the above diffusion models to the image harmonization task faces the significant challenge of enormous computational resource consumption due to training from scratch. For instance, Palette is trained with a batch size of 1024 for 1M steps and SR3+ is trained with a batch size of either 256 or 512 for 1.5M steps. To address this issue, a straightforward approach is to construct an image harmonization model based on an off-the-shelf latent diffusion model~\cite{rombach2022high}. Since the images generated by latent diffusion trained on large-scale datasets are mostly harmonious, the image harmonization model built on top of it can converge quickly.

However, applying a pre-trained latent diffusion model to image harmonization task also faces a significant challenge, which is the reconstruction error caused by the image compression autoencoder. The latent diffusion model takes as its input a feature map of an image that has undergone KL-reg VAE encoding (compressing) process, resulting in a reduced resolution of 1/8 relative to the original image. In other words, if a 256px resolution image and mask are inputted into the latent diffusion model, it will process a feature map and mask with resolution of only 32px. This makes it difficult for the model to reconstruct the content of the image, especially in the case of faces, even if it can generate harmonious images. Jiajie et al. ~\cite{Ji.ji.Ch.ea.2023} tried to build an image harmonization model on the pre-trained Stable Diffusion model, but did not consider this issue, they could only obtain results that were significantly worse than SOTA. 

To address this issue, in this paper, we construct an image harmonization model called DiffHarmony based on a pre-trained latent diffusion model. DiffHarmony tends to generate harmonious but potentially blurry initial images. Therefore, we propose two simple but effective strategies to enhance the clarity of the initially harmonized images. One is to resize the input image to higher resolution to generate images with a higher resolution during inference. The second is to introduce an additional refinement stage that utilizes a simple UNet-structured model to further alleviate the image distortion.

Overall, the main contribution of this work is twofold. First, a method is proposed to enable the pre-trained latent diffusion models to achieve SOTA results on the image harmonization task. Secondly, a wealth of experiments are designed to analyze the advantages and disadvantages of applying the pre-trained latent diffusion models to the image harmonization task, providing a basis for future improvements.

\section{Method}

\begin{figure}[htbp]
    \centering
    \includegraphics[width=1\linewidth]{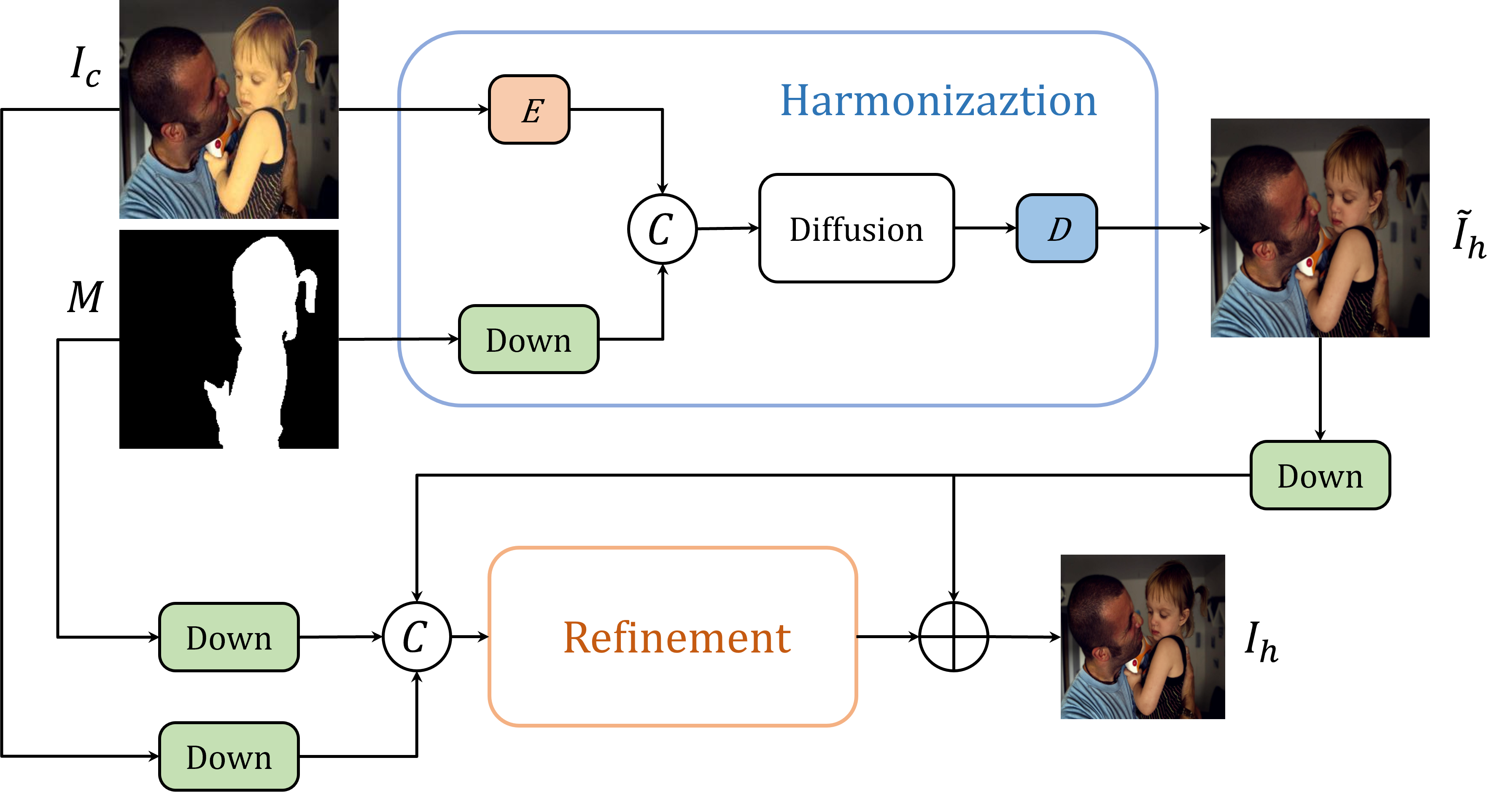}
    \caption{Architecture of our method. In the harmonization stage involving DiffHarmony, composite image $I_c$ and foreground mask $M$ are concatenated as image condition after encoded through VAE and downsample respectively. The diffusion model performs inference, and the output is mapped back to image space through VAE decoder, resulting $\tilde{I}_h$. In the refinement stage, we scale down $\tilde{I}_h$, $I_c$, $M$ and concatenate them together as input to refinement model. After adding refinement model output to downscaled $\tilde{I}_h$, final refined image, $I_h$ is obtained. }
    \label{fig:architecture}
\end{figure}

In this section, we first present the process of modifying a pre-trained latent diffusion model, i.e. Stable Diffusion, to do image harmonization task. Then, we elucidate the techniques to mitigate image distortion issue. The overall architecture of our method is displayed as Figure \textcolor{red}{\ref{fig:architecture}}. 

\subsection{DiffHarmony: Adapting Stable Diffusion}
In typical image harmonization task setup, one needs to input a composite image $I_c$ along with its corresponding foreground mask $M$. Model output is harmonized image $I_h$. Due to this workflow, image harmonization can be categorized as conditional image generation task, thus we can try to utilize pretrained image generation model. Stable Diffusion is the most suitable choice as it's open source,  pretrained on a large amount of diverse data, and already capable of generating images with reasonable content and lighting. However we need to do two adaptations : \textbf{1)} add additional input $I_c$ and $M$ to Stable Diffusion model ; \textbf{2)} use null text input (cause text information is not available in traditional harmonization task).

\subsubsection{\textbf{Inpainting Variation}}
Referring to previous image conditioned diffusion models\cite{saharia2022image, Saharia2021PaletteID,Dhariwal2021Diffusion}, we can extend dimension of the input channel by concatenating image conditions and noisy image input. In image harmonization, the conditions are $I_c$ and $M$. Stable Diffusion inpainting suits our needs. It incorporates additional input channels for masks and masked images and is specifically fine-tuned to do image inpainting task and, same as image harmonization, it generates new foreground content while keeping background part unchanged. 

\subsubsection{\textbf{Null Text Input}}
In the actual generation process, Stable Diffusion typically employs Classifier-Free Guidance (CFG)\cite{ho2021classifier} technique. To perform CFG during inference one needs to train both an unconditional denoising diffusion model $p_\theta(z)$ (parameterized as $\epsilon_\theta(z)$) and a conditional denoising diffusion model $p_\theta(z|c)$ (parameterized as $\epsilon_\theta(z|c)$). In practice, we use a single neural network to incorporate both. For the unconditional part, we can simply input an empty token $\emptyset$, i.e., $\epsilon_\theta(z) = \epsilon_\theta(z , c=\emptyset)$. During inference, we use the formula $\tilde{\epsilon}_\theta(z,c) = (1+w)\cdot\epsilon_\theta(z,c) - w\cdot\epsilon_\theta(z)$ to obtain noise estimations for each step. In image harmonization task, we utilize the unconditional part of Stable Diffusion by inputting only the image conditions while leaving the text empty.

\subsection{Alleviate Image Distortion}
Stable Diffusion uses its VAE encoder to compress image to a lower-resolution upon which the diffusion part does training and inference. The denoised output is mapped back to image space through VAE decoder. When the image resolution is too low, severe image distortion occurs. It can lead to visibly altered object shapes or fluctuations in surface textures. Since image harmonization tasks typically use pixel-level evaluation metrics (e.g., mean squared error), these artifacts can significantly impact the model's overall performance.

\subsubsection{\textbf{Harmonization At Higher Resolution}}
We propose using higher-resolution image inputs for DiffHarmony. In previous work models are typically trained and evaluated at resolution of 256px, but we notice that the image distortion problem becomes excessively severe, which limits the upper bound of image generation quality. Besides, performing inference with Stable Diffusion at 256px does not yield reasonable outputs since it's trained exclusively on 512px images. So we perform inference at 512px or higher resolution. To be consistent with other models in evaluation, we subsequently scale them down to 256px.

\subsubsection{\textbf{Add Refinement Stage}}
To further mitigate the image distortion issue, we introduce an additional refinement stage to enhance the output of DiffHarmony. After harmonization stage, we got $\tilde{I}_h$. Then, the refinement stage makes $\tilde{I}_h$ smoother and repair its texture. We also input $I_c$ and $M$ together because they provide information of texture and shape in uncorrupted image. All inputs are scale down to 256px and concatenated along channel dimension. We introduce skip connection between input $\tilde{I}_h$ and output $I_h$, allowing model to learn the residual instead of outputing refined image directly, which accelerates training convergence. 

\section{Experiment}
\begin{table*}[htbp]
\resizebox{1.0\textwidth}{!}{ 
\begin{tabular}{c|c|ccccccccccccc}
\hline
Dataset & Metric & Composite & DIH\cite{tsai2017deep} & $\text{S}^2\text{AM}$\cite{cun2020improving} & DoveNet\cite{cong2020dovenet} & BargainNet\cite{cong2021bargainnet} & Intrinsic\cite{guo2021intrinsic} & RainNet\cite{ling2021region} & $\text{iS}^2\text{AM}$\cite{sofiiuk2021foreground} & D-HT\cite{guo2021image} & SCS-Co\cite{hang2022scs} & HDNet\cite{Ch.Gu.Li.ea.2023} & Li\cite{Ji.ji.Ch.ea.2023} $et\ al.$ & Ours \\
\hline
\multirow{3}{*}{HCOCO} & PSNR↑ & 33.94 & 34.69 & 35.47 & 35.83 & 37.03 & 37.16 & 37.08 & 39.16 & 38.76 & 39.88 &\textcolor{blue}{41.04}&34.33&\textcolor{red}{41.25}\\
                        & MSE↓ & 69.37 & 51.85 & 41.07 & 36.72 & 24.84 & 24.92 & 29.52 & 16.48 & 16.89 & 13.58&\textcolor{blue}{11.60}&59.55&\textcolor{red}{9.22} \\
                        & fMSE↓ & 996.59 & 798.99 & 542.06 & 551.01 & 397.85 & 416.38 & 501.17 & 266.19 & 299.30 & 245.54& - & - & \textcolor{red}{153.60} \\
\hline
\multirow{3}{*}{HAdobe5k} & PSNR↑ & 28.16 & 32.28 & 33.77 & 34.34 & 35.34 & 35.20 & 36.22 & 38.08 & 36.88 & 38.29 &\textcolor{red}{41.17}&33.18&\textcolor{blue}{40.29}\\
                        & MSE↓ & 345.54 & 92.65 & 63.40 & 52.32 & 39.94 & 43.02 & 43.35 & 21.88 & 38.53 & 21.01&\textcolor{red}{13.58}&161.36&\textcolor{blue}{17.78} \\
                        & fMSE↓ & 2051.61 & 593.03 & 404.62 & 380.39 & 279.66 & 284.21 & 317.55 & 173.96 & 265.11 & 165.48&-&-&\textcolor{red}{107.04} \\
\hline
\multirow{3}{*}{HFlickr} & PSNR↑ & 28.32 & 29.55 & 30.03 & 30.21 & 31.34 & 31.34 & 31.64 & 33.56 & 33.13 & 34.22 &\textcolor{blue}{35.81}&29.21&\textcolor{red}{36.99}\\
                        & MSE↓ & 264.35 & 163.38 & 143.45 & 133.14 & 97.32 & 105.13 & 110.59 & 69.97 & 74.51 & 55.83&\textcolor{blue}{47.39}&224.05&\textcolor{red}{29.68} \\
                        & fMSE↓ & 1574.37 & 1099.13 & 785.65 & 827.03 & 698.40 & 716.60 & 688.40 & 443.65 & 515.45 & 393.72&-&-&\textcolor{red}{199.59} \\
                        
\hline
\multirow{3}{*}{Hday2night} & PSNR↑ & 34.01 & 34.62 & 34.50 & 35.27 & 35.67 & 35.69 & 34.83 & 37.72 & 37.10 & 37.83 &\textcolor{red}{38.85}&34.08&\textcolor{blue}{38.35}\\
                        & MSE↓ & 109.65 & 82.34 & 76.61 & 51.95 & 50.98 & 55.53 & 57.40 & 40.59 & 53.01 & 41.75 &\textcolor{blue}{31.97} & 122.41 & \textcolor{red}{24.94}  \\
                        & fMSE↓ & 1409.98 & 1129.40 & 989.07 &1075.71 & 835.63 & 797.04  & 916.48 & 590.97 & 704.42 & 606.80 &-&-&\textcolor{red}{502.40} \\
\hline
\multirow{3}{*}{Average} & PSNR↑ & 31.63 & 33.41 & 34.35 & 34.76 & 35.88 & 35.90 & 36.12 & 38.19 & 37.55 & 38.75 &\textcolor{red}{40.46}&32.70&\textcolor{blue}{40.44}\\
                        & MSE↓ & 172.47 & 76.77 & 59.67 & 52.33 & 37.82 & 38.71 & 40.29 & 24.44 & 30.30 & 21.33&\textcolor{blue}{16.55}&141.84&\textcolor{red}{14.29} \\
                        & fMSE↓ & 1376.42 & 773.18 & 594.67 & 532.62 & 405.23 & 400.29 & 469.60 & 264.96 & 320.78 & 248.86&-&-&\textcolor{red}{151.42} \\
\hline
\end{tabular}
}
\caption{Quantitative comparison across four sub-datasets of iHarmony4 and in general. Top two performance are shown in \textcolor{red}{red} and \textcolor{blue}{blue}. ↑ means the higher the better, and ↓ means the lower the better. }
\label{tab:result}
\end{table*}

\subsection{Experiment Settings}
\subsubsection{\textbf{Dataset}}
We use iHarmony4\cite{cong2020dovenet} for training and evaluation. iHarmony4 consists of 73,146 image pairs and comprises four subsets: HAdobe5k, HFlickr, HCOCO, and Hday2night. Each sample is composed of a natural image, a foreground mask, and a composite image. Following \cite{cong2020dovenet} , we split the iHarmony4 dataset into training and test sets, containing 65,742 and 7,404 image pairs respectively.

\subsubsection{\textbf{Implementation Detail}}
We trained our DiffHarmony model based on the publicly available Stable Diffusion inpainting model checkpoint on HuggingFace \textcolor{blue}{\footnote{https://huggingface.co/runwayml/stable-diffusion-inpainting}}. We use the Adam optimizer with $\beta_1=0.9$, $\beta_2=0.999$. We employ exponential moving average (EMA) to save model weights, with a decay rate of 0.9999. We use global batch size 32. We initially train the model for 150,000 steps with a learning rate of 1e-5, then reduce the learning rate to 1e-6 and continue our training for additional 50,000 steps. Data augmentations including random resized crop and random horizontal flip are applied. All images are resized to 512px. During training, we use the same noise schedule as the Stable Diffusion model, but use Euler ancestral discrete scheduler~\cite{Ka.Ai.Ai.ea.2022} to generate the samples in only 5 steps during inference.  

Our refinement model is based on the U-Net architecture. $\tilde{I}_h$ are generated at 512px resolution then downscaled to 256px. The harmonization stage can produces diverse results for the same input, which serves as a way of data augmentation during training of the refinement model.

\subsubsection{\textbf{Evaluation}}
In accordance with \cite{cong2021bargainnet,cong2020dovenet,ling2021region}, we use the Peak Signal-to-Noise Ratio (PSNR), Mean Squared Error (MSE), and Foreground MSE (fMSE) metrics on the RGB channels to evaluate the harmonization results. fMSE only calculates the MSE within the foreground regions, providing a measure of foreground harmonization quality.

\subsection{Performance Comparison}
\subsubsection{\textbf{Qualitative Results}}

\begin{figure*}[htb]
    \centering
    \includegraphics[width=1.0\linewidth]{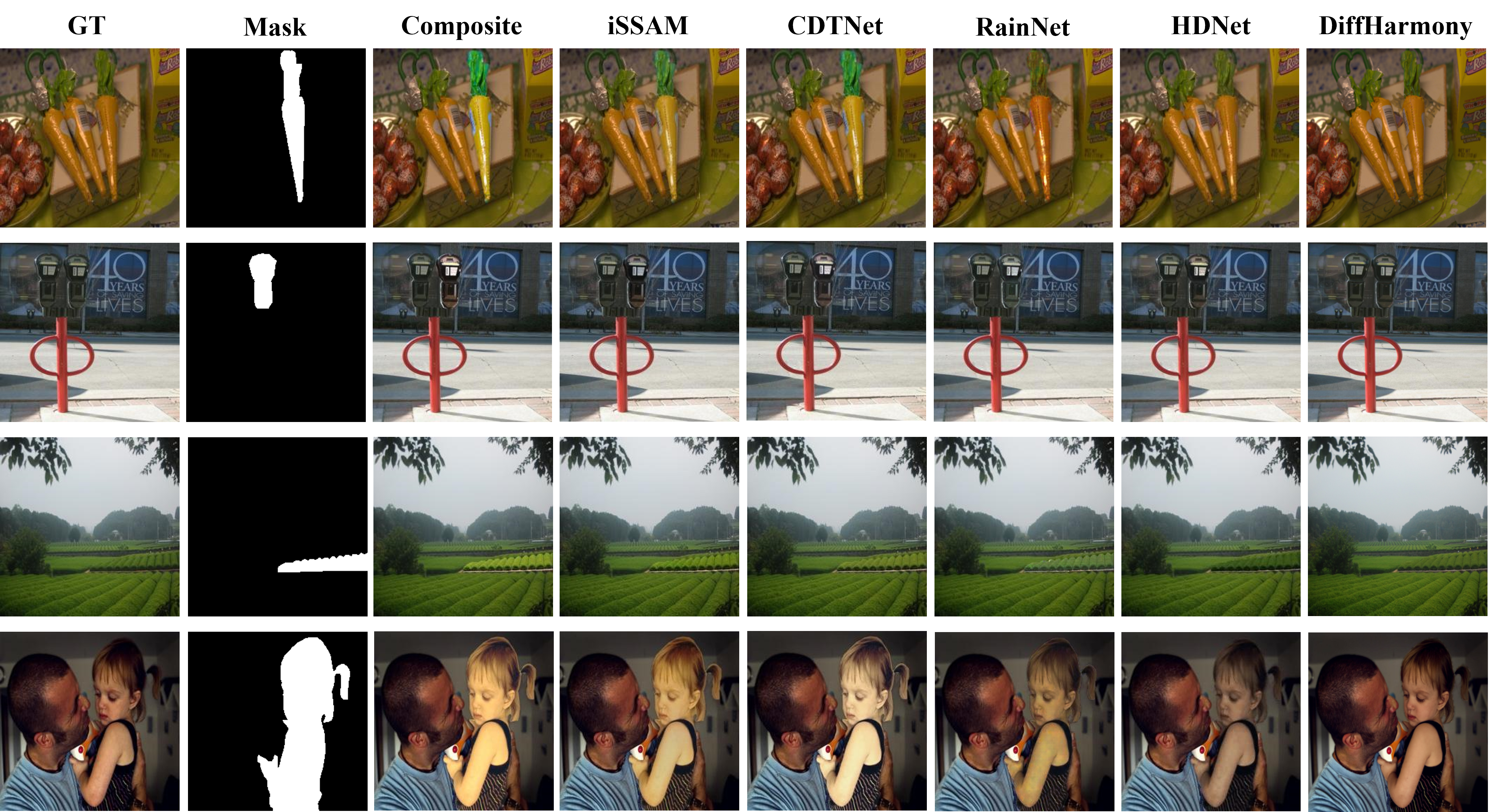}
    \caption{Qualitative comparison on samples from the test set of iHarmony4.}
    \label{fig:harmonization sample}
\end{figure*}

We conduct detailed analysis of model performance and compare qualitatively with previous competing methods. Our method has achieved better visual consistency compared to other approaches as shown in Figure \textcolor{red}{\ref{fig:harmonization sample}}. 

\subsubsection{\textbf{Quantitative Results}}

Table \textcolor{red}{\ref{tab:result}} presents the quantitative results. From Table \textcolor{red}{\ref{tab:result}}, it is evident that our method achieves superior results on most of the sub-datasets. While our method exhibits slightly lower PSNR compared to HDNet, this may be attributed to HDNet using the ground truth background as input during both training and inference. Our method demonstrates significant performance improvements on more challenging subsets HFlickr and Hday2night, indicating gains from pre-trained models for learning in domains with limited data. 

Li $et\ al.$\cite{Ji.ji.Ch.ea.2023} also use Stable Diffusion to do image harmonization task, but they employ a ControlNet-based\cite{Zhang2023AddingCC} approach. As can be seen from Table \textcolor{red}{\ref{tab:result}}, our method is far more advantageous.

\subsection{\textbf{Ablation Study}}

\subsubsection{\textbf{Higher Resolution At Inference}}

\begin{table}[htbp]
    \centering
    \begin{tabular}{c|c|c|c|c}
    \hline
        inf res & refine & PSNR↑ & MSE↓ & fMSE↓ \\
    \hline
        512px & \ding{56} & 37.65 & 26.14 & 290.66 \\
    \hline
        512px & \ding{52} & 39.47 & 19.59 & 205.07 \\
    \hline
        1024px  & \ding{56} & 40.12 & 15.56 & 166.19\\
    \hline
         1024px & \ding{52} & 40.44 & 14.29 & 151.42\\
    \hline
    \end{tabular}
    \caption{Ablation study on using different input resolution and w/wo refinement stage.}
    \label{tab:resolution-refinement}
\end{table}

Table \textcolor{red}{\ref{tab:resolution-refinement}} shows the changes on overall performance when input different resolutions images during harmonization stage. It is obvious that increasing input resolution from 512px to 1024px results in a significant improvement in all metrics, which is reasonable, as higher-resolution input images lead to less information compression.

\subsubsection{\textbf{Refinement Stage}}
We conduct experiments of inference with and without refinement stage. As shown in Table \textcolor{red}{\ref{tab:resolution-refinement}}, adding the refinement stage results in an improvement in the overall performance. The benefit of introducing refinement stage is more prominent when the harmonization stage uses lower image resolutions, as the refinement stage and using higher resolution input both aim to address the issue of image distortion, and they complement each other.

\subsubsection{\textbf{Randomness}}

DiffHarmony in our method is essentially an generative model, but in harmonization task, we usually do not want possible pixel value to vary too much. Therefore, we conduct analysis of randomness. We obtain five groups of results based on five different random seeds, and calculate their mean and std. As shown in Table \textcolor{red}{\ref{tab:randomness}}, the model exhibits small variances, indicating that the harmonization results generated by our method are stable.

\begin{table}[htbp]
    \centering
    \begin{tabular}{c|c|c}
    \hline
        PSNR↑ & MSE↓ & fMSE↓ \\
    \hline
        $37.66\pm 0.02$ & $25.44\pm 0.31$ & $291.03\pm 2.08$ \\
    \hline
    \end{tabular}
    \caption{Randomness analysis. Although essentially a generative model, our method can produce stable harmonized results.}
    \label{tab:randomness}
\end{table}

\subsection{Advanced Analysis}

A noticeable fact is that DiffHarmony uses 512px images during training, while other harmonization models are trained in resolution of 256px. To investigate the impact of this strategy on other models, we select the current state-of-the-art model, HDNet, and train it with 512px images, resulting $\text{HDNet}_{512}$. During test, we use 1024px resolution images as input, then scale harmonization results down to 256px for metric calculation.

\begin{table}[htbp]
    \centering
    \begin{tabular}{c|c|c|c}
    \hline
        Model & $0\%\sim 5\%$ & $5\%\sim 15\%$ & $15\%\sim 100\%$ \\
    \hline
         \multirow{3}{*}{$\text{HDNet}_{512}$} & \textbf{PSNR: 45.64} & \textbf{PSNR: 39.97} & PSNR: 34.59 \\
         & \textbf{MSE: 3.16} & \textbf{MSE: 11.33} & MSE: 47.19 \\
         & \textbf{fMSE: 143.93} & fMSE: 129.87 & fMSE: 152.01\\
    \hline
         \multirow{3}{*}{Ours} & PSNR: 43.28 & PSNR: 39.55 & \textbf{PSNR: 34.80}\\
         & MSE: 4.46 & MSE: 11.90 & \textbf{MSE: 40.47} \\
         & fMSE: 173.10 & \textbf{fMSE: 126.69} & \textbf{fMSE: 128.45} \\
    \hline
    \end{tabular}
    \caption{Comparison between HDNet trained with high-resolution images and our method. $\rm HDNet_{512}$ is trained with 512px images, and the inputs are 1024px images during inference. This is exactly the same as the experimental setting of our method.}
    \label{tab:hdnet-vs-ours}
\end{table}

Our preliminary results show that compared to our method, $\text{HDNet}_{512}$ achieves better PSNR and fMSE but slightly worse MSE. This is counterintuitive. We speculate that our method performs better on samples with larger foreground regions, leading to an overall improvement in MSE. To verify this hypothesis, following HDNet\cite{Ch.Gu.Li.ea.2023}, we divide data into three ranges based on the ratio of the foreground region area and the entire image: $0\%\sim 5\%$, $5\%\sim 15\%$, and $15\%\sim 100\%$. We calculate metrics for each range respectively. Our results, as shown in Table \textcolor{red}{\ref{tab:hdnet-vs-ours}}, reveal that our method is worse than HDNet in the $0\%\sim 5\%$ data range but outperforms it in the $15\%\sim 100\%$ data range. Once again, we emphasize that this arises from the higher information compression loss. However, it’s potential that our method can achieve more advanced results with higher image resolution or using better pre-trained diffusion models.

\section{Conclusion}
In this paper, we propose a solution to achieve SOTA results on image harmonization task based on the Stable Diffusion model. In order to solve the problem of compression loss caused by the VAE in latent diffusion models, we design two effective strategies: utilizing higher-resolution images during inference and incorporating an additional refinement stage. In addition, detailed experimental analysis shows that compared with the SOTA method, our method shows obvious advantages when the foreground area is large enough. 
This is a strong evidence that our model's superior harmonization performance compensates its reconstruction loss, laying a solid foundation for research on image harmonization task using diffusion models.

\bibliographystyle{unsrt}
\bibliography{sample-base}

\end{document}